\documentclass[10pt,twocolumn,letterpaper]{article}

\usepackage{cvpr}              
\definecolor{cvprblue}{rgb}{0.21,0.49,0.74}
\usepackage[pagebackref,breaklinks,colorlinks,allcolors=cvprblue]{hyperref}
\usepackage{multirow}
\usepackage[table]{xcolor}
\usepackage{booktabs}   
\usepackage{diagbox}    
\usepackage{makecell}
\usepackage{threeparttable}
\newcommand{\mname}{MERIT }
\newcommand{\Mname}{MERIT}
\newcommand{\dname}{MDRAW }
\newcommand{\Dname}{MDRAW}
\newcommand{\G}{\mathcal{G}}
\newcommand{\D}{\mathcal{D}}
\newcommand{\E}{\mathcal{E}}
\newcommand{\X}{\mathcal{X}}
\newcommand{\Y}{\mathcal{Y}}

\def\paperID{ } 
\def\confName{CVPR}
\def\confYear{2026}

\title{\Mname: \underline{M}ulti-domain \underline{E}fficient \underline{R}AW \underline{I}mage \underline{T}ranslation}

\author{
Wenjun Huang$^{1}$ \quad Shenghao Fu$^{2}$ \quad Yian Jin$^{1}$ \quad Yang Ni$^{3}$ \quad Ziteng Cui$^{4}$\\  Hanning Chen$^{1}$ \quad Yirui He$^{1}$ \quad Yezi Liu$^{1}$ \quad Sanggeon Yun$^{1}$ \quad SungHeon Jeong$^{1}$ \\Ryozo Masukawa$^{1}$ \quad William Youngwoo Chung$^{1}$ \quad Mohsen Imani$^{1}$\\[0.5em]
$^{1}$University of California, Irvine \quad
$^{2}$University of Pennsylvania\\
$^{3}$Purdue University Northwest \quad
$^{4}$The University of Tokyo\\
{\tt\small m.imani@uci.edu}}
\begin{document}

\twocolumn[{
\maketitle

\begin{center}
\includegraphics[width=0.98\linewidth]{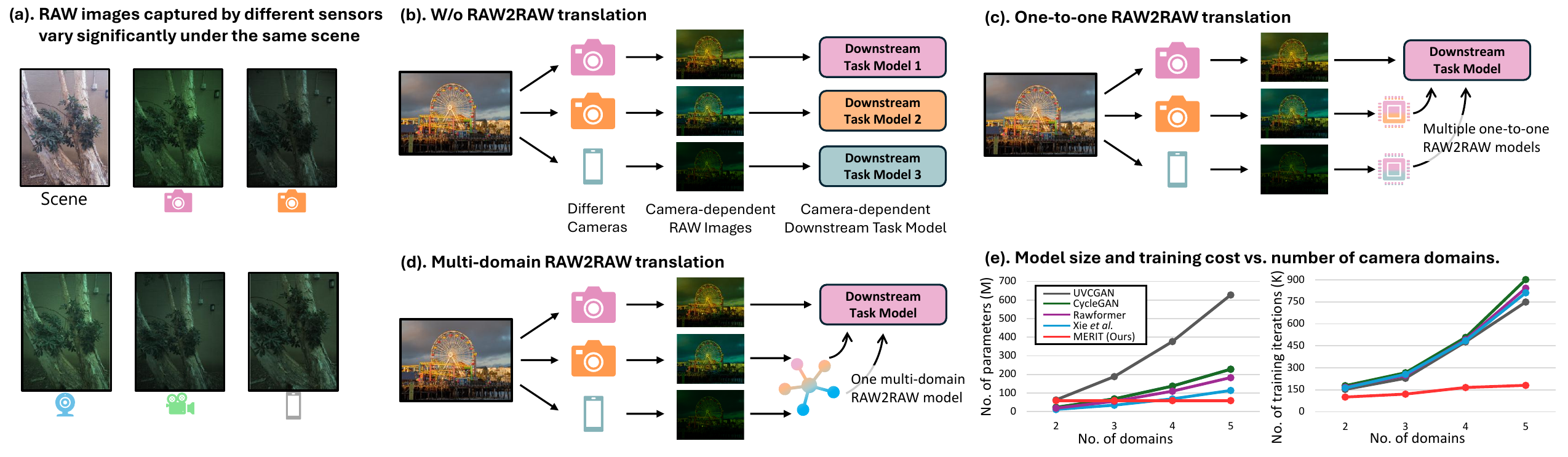}
\vspace{-1em}
\captionof{figure}{\textbf{Overview of the challenges and paradigms in multi-domain RAW-to-RAW (RAW2RAW) translation.}
(a). RAW images captured under the same scene vary across sensors due to differing spectral sensitivity and noise characteristics.
(b). Without RAW2RAW translation, domain-specific task models must be independently trained for each camera.
(c). One-to-one translation enables shared task models, but requires a separate translation model for every source-target pair, leading to scalability issues.
(d). Our proposed method, \Mname, introduces a unified multi-domain RAW2RAW translator that supports all domains with a single model.
(e). \mname achieves superior scalability in both parameter count and training iterations as the number of camera domains increases, outperforming prior methods.
}
\vspace{-1em}
\label{fig:teaser}
\end{center}
\vspace{1em}
}]

\begin{abstract}
RAW images captured by different camera sensors exhibit substantial domain shifts due to varying spectral responses, noise characteristics, and tone behaviors, complicating their direct use in downstream computer vision tasks. 
Prior methods address this problem by training domain-specific RAW-to-RAW translators for each source–target pair, but such approaches do not scale to real-world scenarios involving multiple types of commercial cameras. 
In this work, we introduce \Mname, \textbf{the first unified framework for multi-domain RAW image translation}, which leverages a single model to perform translations across arbitrary camera domains. 
To address domain-specific noise discrepancies, we propose a sensor-aware noise modeling loss that explicitly aligns the signal-dependent noise statistics of the generated images with those of the target domain.
We further enhance the generator with a conditional multi-scale large kernel attention module for improved context and sensor-aware feature modeling.
To facilitate standardized evaluation, we introduce \Dname, \textbf{the first dataset tailored for multi-domain RAW image translation}, comprising both paired and unpaired RAW captures from five diverse camera sensors across a wide range of scenes.
Extensive experiments demonstrate that \mname outperforms prior models in both quality (+5.56 dB) and scalability (80\% reduction in training iterations). Our code is available \href{https://github.com/WJ-Huang/MERIT-Multi-domain-Efficient-RAW-Image-Translation}{here}.
\vspace{-2em}
\end{abstract}

\section{Introduction}
\label{sec:intro}






In recent years, the potential of camera RAW data has been increasingly explored and leveraged across a wide range of downstream vision tasks, including low-level restoration (e.g., super-resolution~\cite{raw_sr_1,raw_sr_2}, low-light imaging~\cite{low-light-raw_SID,low-light-raw_SIED}, and reflection removal~\cite{raw_reflection}), high-level perception (e.g., object detection~\cite{cui2024raw, AdaptiveISP}, instance segmentation~\cite{chen2023instance, huangdr}, and pose estimation~\cite{lee2023human}), and 3D reconstruction~\cite{mildenhall2022rawnerf, jin2024le3d, cui2025luminance}. 
Alongside the growing research interest in RAW-based vision tasks, commercial cameras and smartphones have also been rapidly evolving. For instance, flagship smartphones introduce diverse RAW formats, such as Apple ProRAW\texttrademark, Samsung Expert RAW\texttrademark, and standard DNG options on Android devices~\cite{web_digitaltrends_proraw_2022,web_phonearena_androidraw_2024}, reflecting the growing heterogeneity of RAW data across mobile imaging systems. 
The diversity in RAW formats also leads to significant inter-device variability in captured data, even when capturing the same scene; this variability can significantly affect color, dynamic range, and noise structure (\cref{fig:teaser}~(a)). 
Therefore, multiple camera-specific downstream task models are needed (\cref{fig:teaser}~(b)).
This domain shift roots in the physics of image formation \cite{nguyen2014raw}. An ideal image formation can be expressed as:
{
\setlength{\abovedisplayskip}{2pt}
\setlength{\belowdisplayskip}{2pt}
\begin{equation}
    I(x) = \int_{\omega} R_c(\lambda)S(x,\lambda)L(\lambda)\textbf{d}\lambda,
    \label{eq:raw}
\end{equation}
}
where $\lambda$ represents the wavelength, $\omega$ denotes the visible spectrum, $\rho(x,\lambda)$ is the spectral power distribution of the illuminant. 
The term $S(x,\lambda)$ represents the scene's spectral response at pixel $x$, and $L(\lambda)$ is the lighting in the scene.
$R_c$ is the camera's spectral response, and $c$ is the color channel, which are the variables that change in this case.

Recent work \cite{afifi2021semi, perevozchikov2024rawformer, xie2024generalizing, nikonorov2025color} addressed this challenge through domain-specific RAW-to-RAW (RAW2RAW) translation, which seeks to map RAW images from one camera’s domain to another, enabling the reuse of a fixed downstream task model trained on a particular camera (\cref{fig:teaser}~(c)). 
These methods rely on supervised or unsupervised learning to perform domain alignment, but are inherently limited to one-to-one mappings between two camera domains.
As a result, they do not scale well in scenarios involving multiple camera domains, where training a separate model for each camera pair becomes costly and impractical.

To overcome this scalability bottleneck, we propose the first \textbf{unified framework for \underline{m}ulti-domain \underline{e}fficient \underline{R}AW \underline{i}mage \underline{t}ranslation} (\Mname). 
\mname is capable of translating a RAW image from any source domain to any desired target domain, using a single model conditioned on domain embeddings (\cref{fig:teaser}~(d)). 
\mname introduces a unified multi-domain RAW2RAW translation architecture that enables one-to-many and many-to-many transfers using a single model. 
It maintains a superior translation performance compared with prior work while improving the scalability as the number of domains increases (\cref{fig:teaser}~(e)).

While existing methods attempt to learn RAW domain characteristics implicitly through adversarial training, we observe that a significant portion of domain-specific variation arises from camera-dependent noise patterns. 
Therefore, motivated by the physical Poisson-Gaussian nature of RAW noise~\cite{foi2008practical}, we introduce a noise-aware loss that guides the model to match the statistical noise properties of the target domain. 
This shift from implicit to explicit modeling improves both the realism and fidelity of the translated RAW images, especially in noise-sensitive regions.

In addition, to enable effective context aggregation for RAW2RAW translation, we introduce a novel Multi-Scale Large Kernel Attention (MS-LKA) module that enhances spatial perception and enables domain-adaptive modulation, 
allowing the model to capture fine-grained local patterns and long-range dependencies simultaneously.
This is particularly well-suited for RAW images, where signal distributions and noise characteristics vary across spatial scales and sensor types.
This design not only improves the model’s capacity to learn sensor-specific mappings but also maintains architectural efficiency with minimal parameter increase. 

To facilitate training and evaluation in this new setting, we also introduce \textbf{the first benchmark dataset for \underline{m}ulti-\underline{d}omain \underline{RAW} translation} (named \Dname), composed of 519 unpaired RAW images captured across a diverse set of camera sensors, lighting conditions, and scenes. 
\dname also includes multiple aligned image groups (285 images / 57 groups) across camera domains, enabling quantitative evaluation.
The key contributions of the paper are:
\begin{itemize}
    \item We propose \Mname, the first multi-domain RAW2RAW translation framework that generalizes RAW image translation across multiple camera domains using a single model. Our architecture can flexibly convert RAW images between arbitrary domain pairs during inference.
    \item We introduce a novel explicit noise modeling that enforces statistical consistency of signal-dependent noise across domains, improving translation fidelity under challenging conditions.
    \item We employ a multi-scale large kernel attention to enhance the spatial perception, enabling domain-adaptive modulation given target domain information.
    \item We construct and release \Dname, the first benchmark dataset of multi-domain RAW images for training and evaluating RAW2RAW translation at scale.
    \item Extensive experiments on existing and \dname demonstrate that \mname significantly outperforms prior models in both quality (+5.56 dB) and scalability (2$\times$ smaller).
\end{itemize}

\section{Method}
\label{sec:method}

\begin{figure*}[ht]
  \centering
  \includegraphics[width=\linewidth]{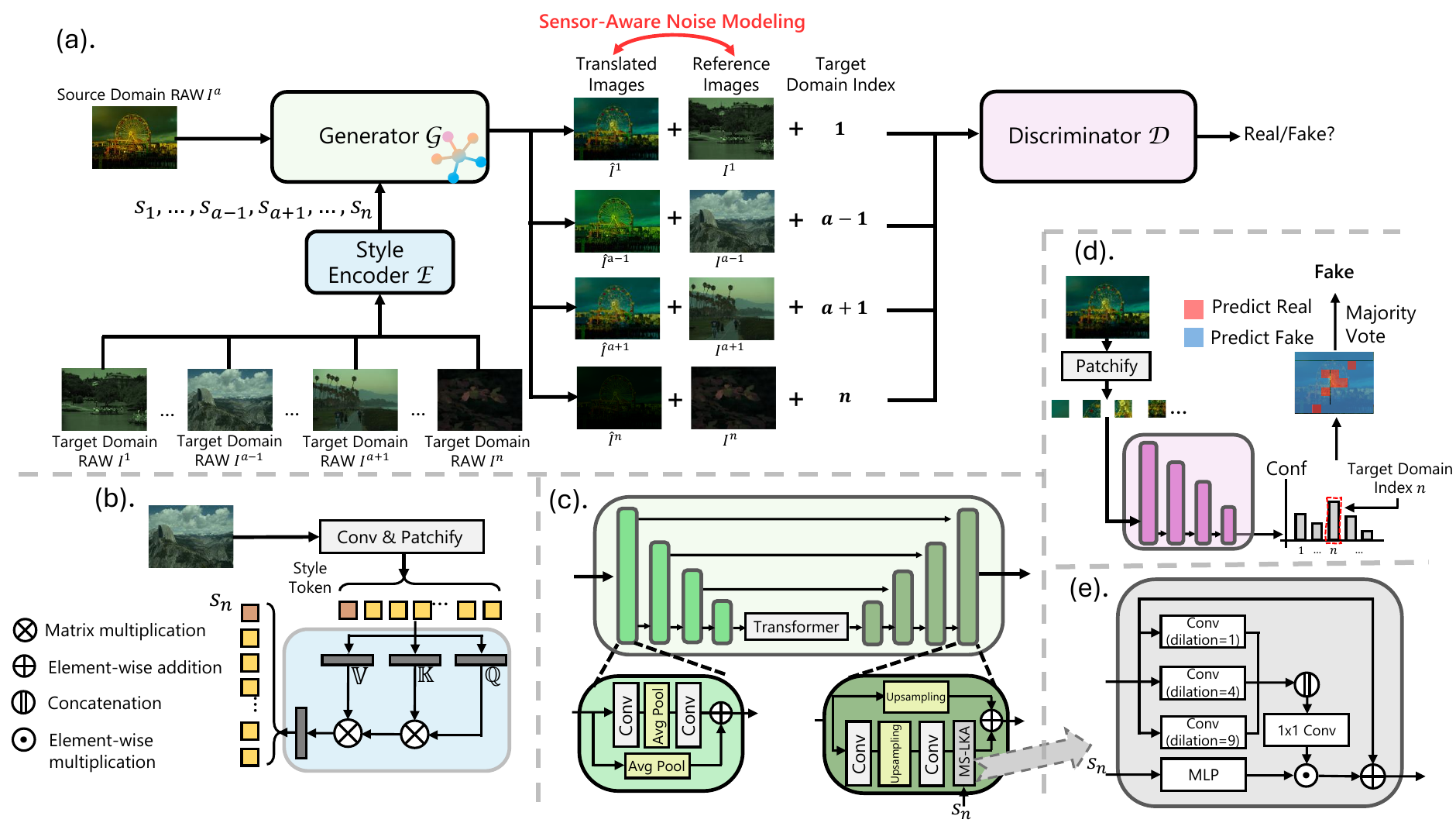}
  \vspace{-1.5em}
   \caption{\textbf{\mname Framework Overview.}
    (a). Overall architecture of \Mname.
    (b). The style encoder 
    $\E$ extracts domain-specific embeddings from target RAW exemplars using a transformer-based architecture.
    (c). The generator $\G$ synthesizes target-domain RAW images.
    (d). The discriminator $\D$ operates on image patches and predicts the realism of each patch, using majority voting to make final decisions.
    (e). The proposed MS-KLA module captures multi-scale features and modulates them via style-aware channel attention.
    }
    \vspace{-1.5em}
    \label{fig:framework}
\end{figure*}

\subsection{Overall Framework}
\label{subsec:overall}
Let $\X$ and $\Y$ be the sets of RAW images and RAW domains, respectively.
Given an image $I^a \in \X$ from the source domain $a\in\Y$, our goal is to train a \textbf{single} generator $\G$ that can generate the corresponding RAW images $\hat{I}^b$ of an arbitrary target domain except the source domain $b\in\Y \setminus\{a\}$.
We use a learnable style encoder $\E$ to generate domain-specific style embedding for each domain and train $\G$ to reflect the style embeddings.
\cref{fig:framework}~(a) illustrates an overview of \Mname, which consists of three key modules.


\noindent\textbf{Generator} \textbf{(}\cref{fig:framework}~(c)\textbf{).} The generator $\G$ translates an input image $I^a$ into an output image $\hat{I}^b = G(I^a, s_b)$ reflecting the domain-specific style embedding $s_b$, which is provided by $\E$. 
$s_b$ is designed to represent the style of a specific domain $b$, which removes the necessity of providing a reference image to $\G$ and allows $\G$ to translate images of all domains. 


\noindent\textbf{Style encoder} \textbf{(}\cref{fig:framework}~(b)\textbf{).} Given an image $I^b$ from domain $b$, our encoder $\E$ extracts the style embedding $s_b = \E(I^b)$ of $I^b$. 
The extracted $s_b$ is domain-specific and content-independent.
This means any images from the same domain should result in a similar embedding, while it may still depend on some characters (e.g., brightness). We discuss this in Sec.~\ref{sec:exp}.


\noindent\textbf{Discriminator} \textbf{(}\cref{fig:framework}~(d)\textbf{).} 
We also incorporate a discriminator $\D$ to supervise the training of $\G$.
$\D$ receives as input the translated image, reference images from the target domain, and the target domain index, and determines whether the input is a real sample from the target domain or a synthesized output.
Following the patch-based adversarial strategy of~\cite{isola2017image}, each image is partitioned into smaller patches, which are independently evaluated by $\D$ to classify each as real or fake. 
The image-level prediction is obtained via majority voting over the patch-level decisions, encouraging $\G$ to produce globally coherent and locally realistic outputs.

\subsection{Sensor-Aware Noise Modeling (SANM)}
\label{subsec:sanm}
\begin{figure}[ht]
  \centering
  \includegraphics[width=.9\linewidth]{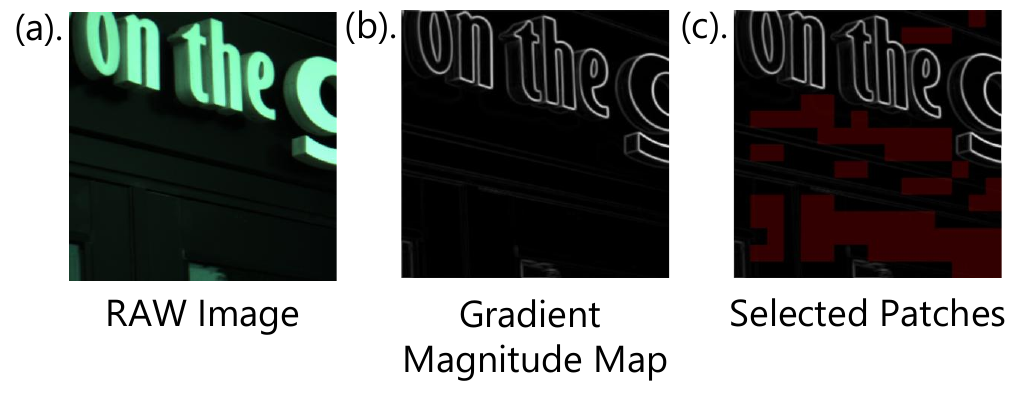}
  \vspace{-1em}
   \caption{\textbf{Visualization of the sensor-aware noise modeling process.} (a). Input RAW image. (b). Gradient magnitude map computed via the Sobel operator. (c). Red-shaded regions indicate low-texture patches selected for sensor noise profile estimation.}
    \label{fig:noise_profile}
    \vspace{-1em}
\end{figure}

Unlike sRGB image-to-image translation, only generating visually realistic images is not sufficient for RAW2RAW translation; capturing the sensor-dependent noise statistics is essential to preserve \textbf{physical plausibility}, especially under high ISO or low-light conditions. 
Traditional GAN-based translation frameworks attempt to learn such characteristics implicitly, but often fail to replicate domain-specific noise patterns accurately.

\noindent \textbf{Noise Modeling in RAW.}
In RAW data, which remains in the linear light domain, the sensor noise can be well-approximated by a Poisson-Gaussian mixture model \cite{foi2008practical}:
{
\setlength{\abovedisplayskip}{2pt}
\setlength{\belowdisplayskip}{2pt}
\begin{align}
    \text{Var}(x)=\alpha\cdot z+\beta,
    \label{eq:poisson_gaussian}
\end{align}
}
where $x$ is the noisy observed signal, $z$ is the true scene intensity, $\alpha$ captures the signal-dependent shot noise, and 
$\beta$ models the signal-independent read noise.
Our goal is to ensure that the translated RAW images exhibit the same noise profile (i.e., the functional dependence between intensity and variance) as the real RAW images from the target sensor domain.
To achieve this, we introduce a differentiable histogram-based noise loss, 
$\mathcal{L}_{noise}$, which aligns the noise characteristics of the generated image with those of the real target-domain RAW images.
Given an input RAW image (\cref{fig:noise_profile}~(a)), we extract small, non-overlapping patches and compute the mean intensity and robust variance estimate via median absolute deviation for each patch. 
To ensure that the variance reflects sensor noise rather than texture, we use a Sobel gradient magnitude filter to identify flat regions, i.e., the patches with minimal structural content. 
(\cref{fig:noise_profile}~(b) visualizes the calculated gradient magnitude map.)
Patches are retained if their average gradient falls below a percentile threshold, as demonstrated in \cref{fig:noise_profile}~(c), making the influence from local texture minimal, and the variance is dominated by sensor noise.

Let $\mu_i$ and $\sigma^2_i$ denote the mean intensity and variance of the $i$-th flat patch.
We bin the patches based on  $\mu_i$ into fixed-width intensity bins (e.g., 100 bins in $[0, 1]$), and compute the average variance per bin.
This yields a noise histogram $\mathcal{H}_{fake}\in\mathbb{R}^{C\times B}$ for the generated image,
where $C$ is the number of channels and $B$ the number of bins.
The target noise profile $\mathcal{H}_{real}$ is precomputed from real images in the target domain using the same procedure and stored as a lookup table. $\mathcal{L}_{noise}$ compares the two histograms:
{
\setlength{\abovedisplayskip}{2pt}
\setlength{\belowdisplayskip}{2pt}
\begin{align}
\mathcal{L}_{noise}=\frac{1}{BC}\sum_{c=1}^C\sum_{b=1}^B|\mathcal{H}_{fake}[c,b]-\mathcal{H}_{real}[c,b]|\cdot \textbf{1}_{\text{valid}}[c,b],
\end{align}
}
where $\textbf{1}_{\text{valid}}$ masks out empty bins to avoid degenerate gradient flow.
$\mathcal{L}_{noise}$ is fully differentiable, robust to image content, and grounded in a physical noise model. 
It enables the generator to learn not only the appearance style of the target sensor domain but also its statistical noise behavior, resulting in more realistic and sensor-faithful outputs.

\subsection{Multi-Scale Large Kernel Attention}
\label{subsec:lka}
RAW images exhibit unique characteristics compared to sRGB data, such as spatially correlated illumination patterns, sensor-specific tone responses, and signal-dependent noise distributions. 
Modeling these requires a network to perceive global spatial relationships while maintaining the precise local structure of the RAW signal. 
Conventional convolutions fail to capture such long-range dependencies, whereas transformer-based attention models are computationally expensive for high-resolution inputs.
To bridge this gap, we extend the concept of large kernel attention~\cite{guo2023visual}, introducing a multi-scale large kernel attention (MS-LKA) module on the upsampling path of $\G$. MS-LKA aggregates features at multiple spatial scales without relying on expensive self-attention, as visualized in \cref{fig:framework}~(e).

\noindent \textbf{Multi-Dilation Feature Extraction.} 
Given an input feature map $F_{in}\in\mathbb{R}^{C\times H\times W}$, we apply three parallel branches of depthwise convolutions with large kernels and distinct dilation rates. 
Each branch captures context at a different receptive field scale, producing intermediate features $F_1, F_2,F_3\in\mathbb{R}^{C\times H\times W}$, which are then concatenated channel-wise and compressed via a $1\times 1$ convolution to maintain dimensionality. 
The resulting fused feature map $F_{concat}=\text{Conv}_{1\times 1}([F_1;F_2;F_3])$ integrates spatial cues across a wide range of receptive fields.

\noindent \textbf{Style-Modulated Channel Attention.} To adaptively weight the multi-scale features according to the target domain, we introduce a style-conditioned attention mechanism.
The style embedding $s$, extracted from the style encoder, is processed by a lightweight feed-forward network (FFN) to produce channel-wise attention weights $A_s\in\mathbb{R}^C$.
These weights are applied to the fused feature map via element-wise multiplication:
$F_{out} = A_s\odot F_{concat}$.

This modulation allows $\G$ to dynamically emphasize domain-relevant channels conditioned on the desired sensor style embedding. 
Compared to conventional convolution blocks or static attention layers, our MS-LKA module offers three distinct advantages for RAW2RAW translation: it expands the effective receptive field while maintaining convolutional inductive bias; it enables domain-aware adaptation through style-guided attention; and it integrates efficiently into our generator without significant parameter overhead.

\subsection{Loss Functions.}
\label{subsec:loss}
Given an image $I^a$, we train our framework using the following objectives. 

\noindent \textbf{Adversarial objective.} 
During training, we sample a target domain $b \in \mathcal{Y} \setminus \{a\} $, and randomly select an RAW image $I^b$ from this domain. 
We compute its style embedding $s_b = \mathcal{E}(I^b)$, and generate a translated image $\hat{I}^b = \mathcal{G}(I^a, s_b)$. 
The discriminator $\mathcal{D}$ is trained to correctly classify real RAW images and distinguish them from generated ones using the adversarial loss:
{
\setlength{\abovedisplayskip}{2pt}
\setlength{\belowdisplayskip}{1pt}
\begin{align}
 \mathcal{L}^\D_{adv} = & \mathbb{E}_{I^a}[\log{\D(a|I^a)}] +   \notag \\
 & \mathbb{E}_{I^a, I^b}[\log{(1 - \D(b|\G(I^a, \E(I^b))))}],
\label{eq:adv_loss}
\end{align}
}
{
\setlength{\abovedisplayskip}{1pt}
\setlength{\belowdisplayskip}{2pt}
\begin{align}
\mathcal{L}_{adv}^\G = 
\mathbb{E}_{I^a, I^b}[\log \mathcal{D}(b|\mathcal{G}(I^a, \mathcal{E}(I^b)))],
\label{eq:adv_gen_loss}
\end{align}
}
where $\D(a|\cdot)$ denotes the output of $\D$ corresponding to domain $a$. $\G$ learns to utilize $s_b$ and generate an image $\G(I^a, s_b)$ that is indistinguishable from real RAW images of the domain $b$. 

\noindent \textbf{Style reconstruction.} 
To enforce $\G$ to utilize the style embedding, we apply a style reconstruction loss:
{
\setlength{\abovedisplayskip}{2pt}
\setlength{\belowdisplayskip}{2pt}
\begin{equation}
\mathcal{L}_{style} = \mathbb{E}_{I^a, I^b} \left[\left\|\mathcal{E}(I^b) - \mathcal{E}(\mathcal{G}(I^a, \mathcal{E}(I^b)))\right\|_1\right]
\label{eq:sty_rec_loss}
\end{equation}
}
Specifically, after generating the image $\hat{I}^b=\G(I^a, \E(I^b))$, we re-extract the style embedding from the generated image using the same encoder $\E$. 
The loss minimizes the L1 distance between the original and reconstructed style embeddings, encouraging style consistency.

\noindent\textbf{Preserving source semantics and content.}
To ensure that $\G$ preserves domain-invariant characteristics, such as content and layout, we adopt a cycle consistency L1 loss:
{
\setlength{\abovedisplayskip}{2pt}
\setlength{\belowdisplayskip}{2pt}
\begin{equation}
\mathcal{L}_{cycle-L1} = \mathbb{E}_{I^a, I^b} \left[ \left\|I^a - \mathcal{G}(\mathcal{G}(I^a, \mathcal{E}(I^b)), \mathcal{E}(I^a))\right\|_1 \right]
\end{equation}
}
Specifically, given a source image $I^a$ and a target style embedding $\E(I^b)$, we first generate the translated image then translate it back to the source domain using the original style embedding $\E(I^a)$, and minimize the L1 distance between the reconstructed image and the original input.

Empirically, unlike sRGB image translation, we observe that solely relying on pixel-wise L1 loss is insufficient for RAW2RAW translation to preserve structural and semantic content. 
The L1 term enforces intensity alignment but fails to maintain consistent texture and fine-grained sensor details. 
\textbf{We introduce a cycle-consistency SSIM loss that complements the L1 reconstruction} by encouraging higher perceptual and structural fidelity between the original and cycle-reconstructed RAW images:
{
\setlength{\abovedisplayskip}{2pt}
\setlength{\belowdisplayskip}{2pt}
\begin{equation}
\mathcal{L}_{\text{cycle-SSIM}} = \mathbb{E}_{I^a, I^b} \left[ 1 - \text{SSIM}\left( I^a, \mathcal{G}(\mathcal{G}(I^a, \mathcal{E}(I^b)), \mathcal{E}(I^a)) \right) \right],
\end{equation}
}
The full loss function is constructed as:
{
\setlength{\abovedisplayskip}{2pt}
\setlength{\belowdisplayskip}{2pt}
\begin{align}
    \mathcal{L}_{total} = & \lambda_1 \mathcal{L}_{noise} + \lambda_2\mathcal{L}_{adv}^\D + \lambda_3\mathcal{L}_{adv}^\G \notag \\ &+ \lambda_4\mathcal{L}_{cycle-L1} +\lambda_5\mathcal{L}_{cycle-SSIM},
\end{align}
}
where $\lambda_1, \cdots, \lambda_5$ are emperical hyper-parameters.
\vspace{-1em}
\section{Experiments}
\label{sec:exp}
\begin{figure}[t]
  \centering
  \includegraphics[width=.7\linewidth]{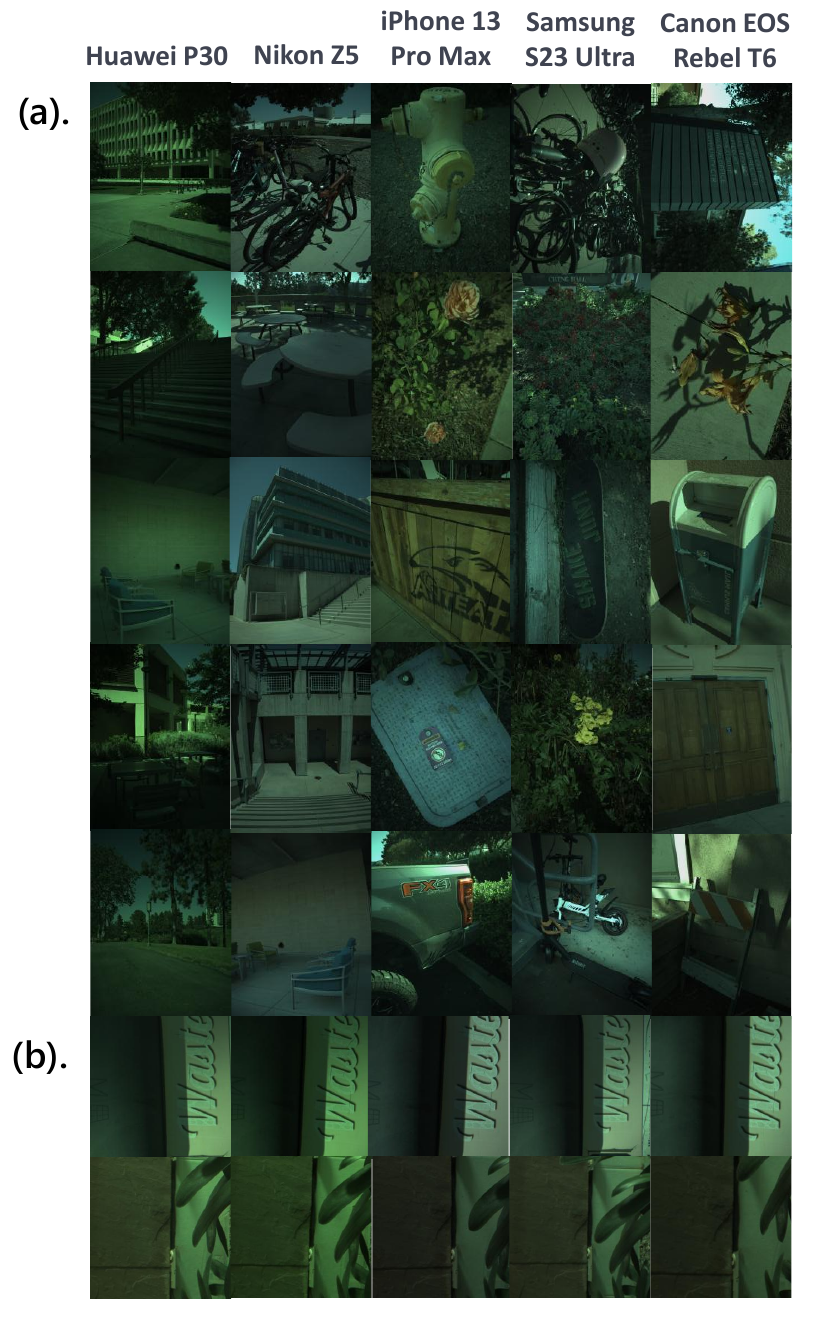}
  \vspace{-1em}
   \caption{
   \textbf{Sample images from our \dname benchmark.} (a). RAW images captured under various lighting and scenes from five different camera sensors. (b). Example of aligned multi-domain RAW captures of the same scene, used for evaluation in cross-domain RAW2RAW translation.
   }
    \label{fig:dataset_visualization}
\end{figure}

\noindent \textbf{Datasets.} In this section, we provide a description of the datasets used in this work.
Particularly, we use two datasets: \textbf{(i).} RAW-to-RAW mapping dataset \cite{afifi2021semi}. It contains a total of 392 unpaired RAW images (196 from each of the Samsung Galaxy S9 and iPhone X). Additionally, the dataset includes 115 paired testing RAW images from each camera for evaluation. \textbf{(ii).} our collected dataset, Multi-Domain RAW (\Dname). 
As one of our contributions, we propose \Dname, a new dataset of RAW images captured by five cameras with different sensors: Samsung Galaxy S23 Ultra, Huawei P30, iPhone 13 Pro, Nikon Z5, and Canon EOS Rebel T6.
\cref{fig:dataset_visualization}~(a) shows example RAW images from each camera. 
To the best of our knowledge, there is no publicly available dataset of RAW images captured by multiple cameras that meets our setup requirements (i.e., containing both unpaired and paired RAW image sets under diverse illuminations and scenes). 
\dname consists of unpaired and paired RAW images for each camera with their corresponding sRGB images.
\cref{tab:camera_sensors} summarizes the sensor and statistics of \Dname.
More details about the datasets are provided in the supplementary material.


To construct a high-quality, pixel-level evaluation benchmark across multiple RAW domains, we also extract spatially aligned patch pairs from images of the same scene taken by different devices (as shown in \cref{fig:dataset_visualization}~(b)).
Since no ground truth pixel-level alignment exists between images from different sensors, we adopt a feature-based correspondence approach to identify regions suitable for evaluation.
Specifically, we extend the LoFTR~\cite{sun2021loftr} to operate across domains, using a multi-stage pipeline that combines dense matching, geometric verification, spatial filtering, and patch-level synchronization. 
The whole construction process is discussed in detail in the supplementary material.

\begin{table}[t]
\vspace{-1em}
\centering
\resizebox{\linewidth}{!}{
\begin{tabular}{lcc}
\toprule\toprule
\textbf{Camera} & \textbf{Sensor} & \textbf{No. of unpaired images} \\
\midrule\midrule
Samsung Galaxy S23 Ultra & ISOCELL HP2 & 81 \\
Huawei P30 & Sony IMX650 Exmor RS & 119 \\
iPhone 13 Pro Max & Sony IMX703 & 98 \\
Nikon Z5 & Sony IMX128 & 114 \\
Canon EOS Rebel T6 & DIGIC 4+ & 107 \\
\bottomrule
\bottomrule
\end{tabular}
}
\vspace{-1em}
\caption{\textbf{Summarization of our collected dataset (\Dname).}}
\vspace{-2em}
\label{tab:camera_sensors}
\end{table}

\noindent \textbf{Metrics.} We assess the models utilizing three commonly accepted metrics: mean absolute error (MAE), peak signal-to-noise ratio (PSNR), structural similarity index measure (SSIM)~\cite{wang2004image}, and KL divergence.
Details about the adopted metrics are discussed in the supplementary material.

\noindent \textbf{Training Details.}
We train our model with a batch size of 8 using the Adam optimizer ($\beta_{1}=0.9$, $\beta_{2}=0.99$), on a NVIDIA H200 GPU. 
The learning rate is fixed at $1\times10^{-4}$ for all network components, including the generator, discriminator, and style encoder. 
The training process is conducted for 200,000 iterations. 
We apply random horizontal flipping ($p=0.5$) as the data augmentation strategy. 
All input RAW images are pre-processed into $256\times256$ patches. 
All models are trained on unpaired images and evaluated on paired image sets to assess cross-domain translation performance.
During evaluation, we adopt a brightness-based reference image selection strategy.
Specifically, for each input, we select the target-domain image from the trainset with the most similar brightness as the reference for style embedding extraction.
Brightness similarity is quantified using channel-wise means: we compute the spatial mean of each of the four Bayer channels to form a 4-dimensional channel-mean vector, which is then used for comparison.
In addition to these settings, we use the following loss weights in our full objective: $\lambda_{\text{reg}}=1$, $\lambda_{\text{cyc}}=10$, $\lambda_{\text{sty}}=1$, $\lambda_{\text{ds}}=5$, $\lambda_{\text{id}}=1.0$, $\lambda_{\text{noise}}=1.0$, and $\lambda_{\text{cyc}}^{\text{SSIM}}=0.1$.


\begin{figure*}[ht]
  \centering
  \includegraphics[width=\linewidth]{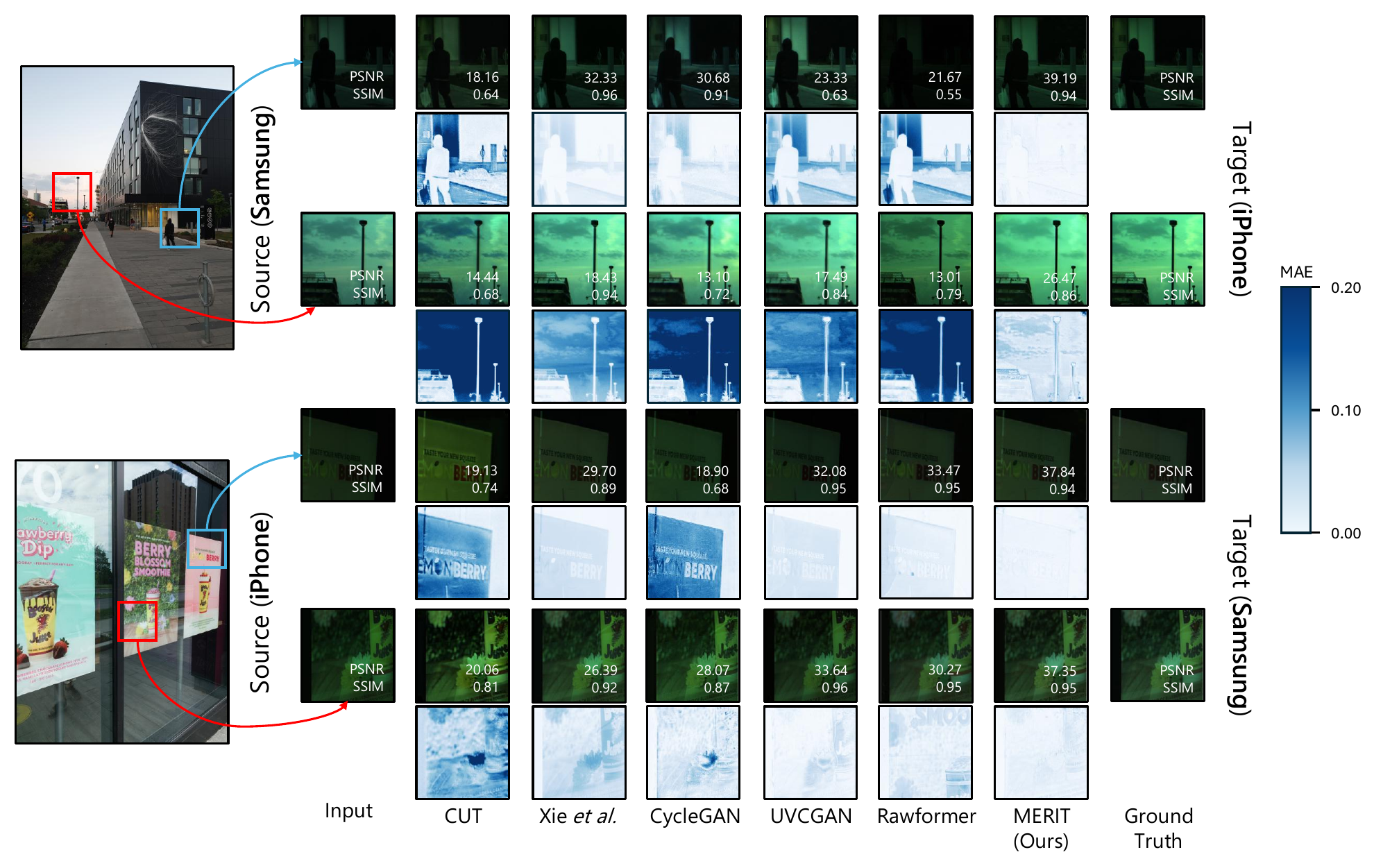}
  \vspace{-2em}
   \caption{\textbf{Qualitative results on the RAW-to-RAW mapping dataset.} Each group of images (from left to right) shows: the input source RAW image, predictions from CUT~\cite{park2020contrastive}, Xie~\etal~\cite{xie2024generalizing}, CycleGAN~\cite{zhu2017unpaired}, UVCGAN~\cite{torbunov2023uvcgan}, Rawformer~\cite{perevozchikov2024rawformer}, our method \Mname, and the target ground truth. Odd-numbered rows display the translated RAW outputs, while even-numbered rows visualize the corresponding absolute error maps with respect to the ground truth.}
    \label{fig:visualization1}
    \vspace{-1em}
\end{figure*}

\subsection{Results}
\label{subsec:results}
\cref{fig:visualization1} presents a visual comparison of different methods on RAW-to-RAW mapping dataset. 
Odd-numbered rows depict the RAW outputs, while even-numbered rows display the corresponding absolute error maps with respect to the ground truth. 
\mname consistently produces translated images with fewer perceptual artifacts and lower reconstruction errors.
Compared to prior methods, \mname generates outputs with more faithful scene illumination and finer structural alignment, especially in regions with sharp edges or strong color contrast. 
The error maps show that \mname leads to smaller residuals, in terms of magnitude and spatial extent, indicating superior pixel-level accuracy. 

\begin{table}[t]
  \centering
  \resizebox{\linewidth}{!}{
  \begin{threeparttable}
    \begin{tabular}{c|c|ccc|ccc}
      \toprule\toprule
      \multirow{2}{*}{Training Paradigm} & \multirow{2}{*}{Model}
        & \multicolumn{3}{c}{Samsung-S9$\to$iPhone-X}
        & \multicolumn{3}{c}{iPhone-X$\to$Samsung-S9} \\
      \cline{3-5} \cline{6-8}
      &                                   
        & PSNR$\uparrow$ & SSIM$\uparrow$ & MAE$\downarrow$  
        & PSNR$\uparrow$ & SSIM$\uparrow$ & MAE$\downarrow$ \\
      \midrule\midrule
      \multirow{3}{*}{Non‐learning}
        & Global calibration (3$\times$3) \cite{nguyen2014raw}  & 24.52 & 0.71 & 0.049 & 17.03 & 0.51 & 0.160 \\
        & Global calibration (poly)      \cite{nguyen2014raw}  & \cellcolor{gray!20}24.88 & \cellcolor{gray!20}0.72 & \cellcolor{gray!20}0.048 & \cellcolor{gray!20}16.88 & \cellcolor{gray!20}0.50 & \cellcolor{gray!20}0.160 \\
        & FDA                             \cite{yang2020fda}  & 20.95 & 0.48 & 0.060 & 19.18 & 0.47 & 0.090 \\
      \midrule
      Semi‐supervised
        & Afifi \emph{et al.}            \cite{afifi2021semi}     & \cellcolor{gray!20}29.65 & \cellcolor{gray!20}0.89 & \cellcolor{gray!20}0.027 & \cellcolor{gray!20}28.58 & \cellcolor{gray!20}0.90 & \cellcolor{gray!20}0.033 \\
      \midrule
      \multirow{6}{*}{Unsupervised}
        & CycleGAN                        \cite{zhu2017unpaired}   & 24.55 & 0.76 & 0.046 & 25.21 & 0.76 & 0.042 \\
        & Cut                             \cite{park2020contrastive}                & \cellcolor{gray!20}23.51 & \cellcolor{gray!20}0.71 & \cellcolor{gray!20}0.050 & \cellcolor{gray!20}22.44 & \cellcolor{gray!20}0.71 & \cellcolor{gray!20}0.053 \\
        & Swin-UNIT                       \cite{li2023swin}                & 23.92 & 0.72 & 0.057 & 23.77 & 0.75 & 0.051 \\
        & Chai \emph{et al.}              \cite{chai2020supervised}                 & \cellcolor{gray!20}29.35 & \cellcolor{gray!20}0.86 & \cellcolor{gray!20}0.028 & \cellcolor{gray!20}27.78 & \cellcolor{gray!20}0.86 & \cellcolor{gray!20}0.037 \\
        & UVCGAN                          \cite{torbunov2023uvcgan}                & 27.22 & 0.82 & 0.031 & 26.10 & 0.79 & 0.037 \\
        & Xie \emph{et al.}                           \cite{xie2024generalizing}             & \cellcolor{gray!20}\underline{29.73} & \cellcolor{gray!20}\underline{0.90} & \cellcolor{gray!20}0.025 & \cellcolor{gray!20}28.09 & \cellcolor{gray!20}\textbf{0.89} & \cellcolor{gray!20}0.037 \\
        & Rawformer                       \cite{perevozchikov2024rawformer}$^{\ast}$         & 29.32 & \underline{0.90} & \underline{0.023} & \underline{28.45} & \underline{0.88} & \underline{0.034} \\
        & \mname(Ours)                            & \cellcolor{gray!50}\textbf{35.29} & \cellcolor{gray!50}\textbf{0.91} & \cellcolor{gray!50}\textbf{0.015}
                                                      & \cellcolor{gray!50}\textbf{31.90} & \cellcolor{gray!50}\underline{0.85} & \cellcolor{gray!50}\textbf{0.021} \\
      \bottomrule\bottomrule
    \end{tabular}%
    \begin{tablenotes}[leftmargin=0.35cm, itemindent=.0cm, itemsep=0.0cm, topsep=0.1cm]
    \item[$\ast$] We retrain the model under the same settings as our \mname and other baselines, since the results reported in the original paper were trained on the images demosaiced using the Menon algorithm~\cite{menon2006demosaicing}.
    \end{tablenotes}
    \end{threeparttable}
  }
  \vspace{-1em}
\caption{\textbf{Quantitative results on the RAW‐to‐RAW mapping dataset.} Best results are \textbf{bolded} and second-best are \underline{underlined}.}
\vspace{-2em}
\label{tab:single_mapping}
\end{table}

\begin{table*}[h!]
\centering
\resizebox{\linewidth}{!}{
\begin{tabular}{c|c|c|c|c|c}
\toprule\toprule
\textbf{\diagbox{Source}{Target}} & \textbf{Samsung} & \textbf{Huawei} & \textbf{iPhone} & \textbf{Nikon} & \textbf{Canon} \\
\midrule
\textbf{Samsung} & \textbf{--} & 
\makecell{ 
\underline{0.026} / 30.77 / \underline{0.77} / 1.53\\
\cellcolor{gray!20}\underline{0.026} / \underline{31.15} / \textbf{0.78} / \underline{1.46} \\
\cellcolor{gray!50}\textbf{0.025} / \textbf{31.27} / \textbf{0.78} / \textbf{1.24}
} 
& \makecell{ 
\underline{0.036} / \textbf{29.11} / 0.75 / \textbf{1.64}\\
\cellcolor{gray!20}\underline{0.036} / 28.88 / \underline{0.76} / 1.89\\
\cellcolor{gray!50}\textbf{0.035} / \textbf{29.11} / \textbf{0.77} / \underline{1.73}
}
& \makecell{ 
0.038 / 27.99 / \underline{0.72} / 2.36\\
\cellcolor{gray!20}\underline{0.037} / \underline{28.21} / \underline{0.72} / \underline{2.28}\\
\cellcolor{gray!50}\textbf{0.034} / \textbf{28.71} / \textbf{0.73} / \textbf{2.06}
} 
& \makecell{ 
0.047 / 26.71 / 0.72 / 3.18\\
\cellcolor{gray!20}\underline{0.045} / \underline{27.16} / \underline{0.73} / \underline{2.77}\\
\cellcolor{gray!50}\textbf{0.043} / \textbf{27.59} / \textbf{0.75} / \textbf{2.65}
} 
\\
\midrule
\textbf{Huawei}  
& \makecell{ 
                                        \textbf{0.028} / \textbf{30.26} / \underline{0.74} / 1.73\\
\cellcolor{gray!20}0.032 / \underline{29.15} / 0.73 / \underline{1.65}\\
\cellcolor{gray!50}\underline{0.029} / \textbf{30.26} / \textbf{0.76} / \textbf{1.51}
}
& -- 
&   \makecell{ 
                                        \textbf{0.033} / \underline{29.20} / \underline{0.77} / 2.38\\
\cellcolor{gray!20}\underline{0.035} / 28.48 / 0.76 / \textbf{2.03}\\
\cellcolor{gray!50}\textbf{0.033} / \textbf{29.23} / \textbf{0.79} / \underline{2.30}
}
&   \makecell{ 
                                        \underline{0.033} / \underline{29.43} / \textbf{0.75} / 2.05\\
\cellcolor{gray!20}\textbf{0.032} / \textbf{29.52} / \underline{0.74} / \underline{1.93}\\
\cellcolor{gray!50}\textbf{0.032} / 29.39 / \textbf{0.75} / \textbf{1.89}
}
& \makecell{ 
                                        0.045 / 26.98 / 0.72 / 3.63\\
\cellcolor{gray!20}\underline{0.043} / \underline{27.46} / \underline{0.73} / \textbf{2.59}\\
\cellcolor{gray!50}\textbf{0.042} / \textbf{27.60} / \textbf{0.75} / \underline{3.07}
}
\\
\midrule
\textbf{iPhone}  
& \makecell{ 
                                        0.038 / 28.71 / \underline{0.73} / \underline{1.55}\\
\cellcolor{gray!20}\underline{0.035} / \underline{29.07} / 0.72 / 1.67\\
\cellcolor{gray!50}\textbf{0.034} / \textbf{29.12} / \textbf{0.74} / \textbf{1.42}
} 
&  \makecell{ 
                                        0.033 / 29.07 / \underline{0.74} / 2.17\\
\cellcolor{gray!20}\underline{0.030} / \textbf{30.19} / \textbf{0.76 }/ \underline{1.91}\\
\cellcolor{gray!50}\textbf{0.029} / \underline{29.77} / \textbf{0.76} / \textbf{1.87}
} 
& -- 
&  \makecell{ 
                                        0.044 / \underline{26.67} / \underline{0.69} / 2.79\\
\cellcolor{gray!20}\textbf{0.039} / \textbf{27.68} / \underline{0.71} / \underline{2.58}\\
\cellcolor{gray!50}\underline{0.038} / \textbf{27.68} / \textbf{0.72} / \textbf{2.50}
}
&  \makecell{ 
                                        0.046 / 26.01 / 0.70 / 2.85\\
\cellcolor{gray!20}\textbf{0.038} / \textbf{28.17} / \underline{0.73} / \textbf{2.30}\\
\cellcolor{gray!50}\underline{0.041} / \underline{27.09} / \textbf{0.74} / \underline{2.47}
}\\
\midrule
\textbf{Nikon}   
& \makecell{ 
                                        \textbf{0.039} / \textbf{27.64} / \underline{0.69} / \underline{2.65}\\
\cellcolor{gray!20}\underline{0.041} / 26.79 / 0.67 / 2.69\\
\cellcolor{gray!50}\textbf{0.039} / \underline{27.37} / \textbf{0.70} / \textbf{2.60}
}
& \makecell{ 
                                        \underline{0.034} / 28.58 / \underline{0.70} / \underline{2.54}\\
\cellcolor{gray!20}\textbf{0.033} / \underline{28.83} / \textbf{0.74} / 2.65\\
\cellcolor{gray!50}\textbf{0.033} / \textbf{28.92} / \textbf{0.74} / \textbf{2.47}
}
&  \makecell{ 
                                        0.044 / \underline{26.54} / \underline{0.71} / \underline{3.15}\\
\cellcolor{gray!20}\textbf{0.043} / 26.32 / \underline{0.71} / \textbf{2.65}\\
\cellcolor{gray!50}\textbf{0.043} / \textbf{26.61} / \textbf{0.73} / 3.16
}
& -- 
& \makecell{ 
                                        \underline{0.033} / 29.31 / 0.76 / \textbf{2.12}\\
\cellcolor{gray!20}\underline{0.033} / \underline{29.38} / \underline{0.77} / \underline{2.20}\\
\cellcolor{gray!50}\textbf{0.032} / \textbf{29.51} / \textbf{0.78} / 2.32
}
\\
\midrule
\textbf{Canon}   
& \makecell{ 
                                        \underline{0.047} / \textbf{26.61} / \textbf{0.70} / 3.06\\
\cellcolor{gray!20}0.049 / 25.78 / 0.66 / \underline{2.82}\\
\cellcolor{gray!50}\textbf{0.045} / \underline{26.49} / \textbf{0.70} / \textbf{2.79}
}
&  \makecell{ 
                                        0.044 / 26.88 / \underline{0.71} / \textbf{3.20}\\
\cellcolor{gray!20}\underline{0.042} / \underline{27.08} / 0.70 / 3.28\\
\cellcolor{gray!50}\textbf{0.041} / \textbf{27.15} / \textbf{0.72} / \underline{3.24}
}
&  \makecell{ 
                                        0.044 / \underline{26.40} / \textbf{0.73} / 2.94\\
\cellcolor{gray!20}\underline{0.043} / 26.07 / \underline{0.71} / \textbf{2.31}\\
\cellcolor{gray!50}\textbf{0.042} / \textbf{26.42} / \textbf{0.73} / \underline{2.63}
}
&  \makecell{ 
                                        0.033 / 29.13 / \underline{0.75} / \underline{2.11}\\
\cellcolor{gray!20}\underline{0.032} / \textbf{29.31} / \underline{0.75} / \textbf{1.99}\\
\cellcolor{gray!50}\textbf{0.031} / \underline{29.27} / \textbf{0.76} / 2.27
}
& --  
\\
\bottomrule\bottomrule
\end{tabular}
}
\caption{\textbf{Cross-domain RAW2RAW translation results on MDRAW.}
Each cell reports results for translation from a source domain (column) to a target domain (row).
Within each cell, three lines correspond to different methods: UVCGAN~\cite{torbunov2023uvcgan}, \colorbox{gray!20}{Xie \emph{et al.}~\cite{xie2024generalizing}}, and \colorbox{gray!50}{MERIT (Ours)}.
Each line contains four metrics in the order of \textbf{MAE}($\downarrow$)/ \textbf{PSNR}($\uparrow$)/ \textbf{SSIM} ($\uparrow$)/ \textbf{KL Divergence}($\downarrow$). }
\label{tab:cross_domain}
\end{table*}

\cref{tab:single_mapping} presents a comprehensive quantitative comparison across various training paradigms. 
\mname consistently outperforms all baselines, achieving state-of-the-art (SOTA) results under the unsupervised setting. 
Traditional non-learning approaches perform poorly, especially in the direction iPhone$\rightarrow$Samsung, with MAE values exceeding 0.16. This demonstrates the inherent limitations of linear mappings in modeling complex inter-sensor differences. 
Afifi~\etal~\cite{afifi2021semi} performs competitively but requires paired supervision. 
In contrast, \mname achieves outstanding results without any paired supervision, validating the strength of our unsupervised multi-domain training strategy.
Among unsupervised baselines, prior GAN-based and contrastive methods yield moderate performance but lag behind recent domain-specific designs, such as Rawformer~\cite{perevozchikov2024rawformer} and Xie~\etal~\cite{xie2024generalizing}.
Even so, \textbf{\mname consistently outperforms these tailored approaches in 5 out of 6 evaluation metrics across both translation directions. }
\mname achieves a PSNR gain of $+5.56$ dB and a MAE reduction of $0.008$ on Samsung$\rightarrow$iPhone, while also improving iPhone$\rightarrow$Samsung by $+3.45$ dB in PSNR and reducing MAE by $0.013$. 
These results emphasize the importance of our innovations in effectively bridging the spectral diversity of real-world cameras, confirming the effectiveness and robustness of \mname in producing high-quality and semantically consistent RAW image translations.

\cref{tab:cross_domain} presents a comprehensive evaluation of our \mname against two strong baselines, UVCGAN~\cite{torbunov2023uvcgan} and Xie \etal~\cite{xie2024generalizing}, across all pairwise source-to-target camera domain combinations in our proposed \dname dataset. 
Each cell summarizes the performance for a specific domain pair using four metrics: MAE, PSNR, SSIM, and KL divergence. 
Across the 20 non-diagonal translation pairs, \mname consistently achieves the best or second-best performance in nearly every metric and direction. 
Specifically, \mname records the \textbf{lowest MAE in 17 out of 20 cases}, the \textbf{highest PSNR in 14 out of 20}, the \textbf{highest SSIM in 20 out of 20}, and the lowest KL divergence in 11 out of 20, demonstrating superior accuracy and perceptual quality in modeling inter-domain RAW distributions.
These results underscore the robustness and generalizability of \mname in diverse cross-camera scenarios, outperforming prior RAW translation models.
\vspace{-1em}
\subsection{Ablation Study}
\label{subsec:ablation}
\begin{table}[t]
\centering
\resizebox{.9\columnwidth}{!}{
\begin{tabular}{llccc}
\toprule\toprule
\textbf{Setting} & \textbf{Source→Target} & \textbf{MAE}$\downarrow$ & \textbf{PSNR}$\uparrow$ & \textbf{SSIM}$\uparrow$ \\
\midrule\midrule
\multicolumn{5}{l}{\textit{Baseline}} \\
\cmidrule(l){1-5}
 & iPhone-X $\rightarrow$ Samsung-S9 & 0.0256 & 30.99 & 0.8357 \\
 & Samsung-S9 $\rightarrow$ iPhone-X & \cellcolor{gray!20}0.0186 & \cellcolor{gray!20}33.75 & \cellcolor{gray!20}0.8841 \\
 & \textbf{Avg all} & \cellcolor{gray!50}0.0221 & \cellcolor{gray!50}32.37 & \cellcolor{gray!50}0.8599 \\
\multicolumn{5}{l}{\textit{+ SANM}} \\
\cmidrule(l){1-5}
 & iPhone-X $\rightarrow$ Samsung-S9 & 0.0250 & 31.24 & 0.8426 \\
 & Samsung-S9 $\rightarrow$ iPhone-X & \cellcolor{gray!20}0.0169 & \cellcolor{gray!20}34.37 & \cellcolor{gray!20}0.8966 \\
 & \textbf{Avg all} & \cellcolor{gray!50}0.0210 & \cellcolor{gray!50}32.71 & \cellcolor{gray!50}0.8682 \\
\addlinespace
    \multicolumn{5}{l}{\textit{+ MS-LKA + SANM}} \\
\cmidrule(l){1-5}
 & iPhone-X $\rightarrow$ Samsung-S9 & 0.0226 & 31.74 & 0.8482 \\
 & Samsung-S9 $\rightarrow$ iPhone-X & \cellcolor{gray!20}0.0164 & \cellcolor{gray!20}34.66 & \cellcolor{gray!20}0.9006 \\
 & \textbf{Avg all} & \cellcolor{gray!50}0.0195 & \cellcolor{gray!50}33.20 & \cellcolor{gray!50}0.8744 \\
\addlinespace
\multicolumn{5}{l}{\textit{+ MS-LKA + SANM + $\mathcal{L}_{cycle-SSIM}$}} \\
\cmidrule(l){1-5}
 & iPhone-X $\rightarrow$ Samsung-S9 & 0.0219 & 31.90 & 0.8508 \\
 & Samsung-S9 $\rightarrow$ iPhone-X & \cellcolor{gray!20}0.0151 & \cellcolor{gray!20}35.29 & \cellcolor{gray!20}0.9104 \\
 & \textbf{Avg all} & \cellcolor{gray!50}\textbf{0.0185} & \cellcolor{gray!50}\textbf{33.60} & \cellcolor{gray!50}\textbf{0.8806} \\
\bottomrule\bottomrule
\end{tabular}
}
\caption{\textbf{Ablation study on RAW-to-RAW mapping dataset.}}
\vspace{-2em}
\label{tab:ablation}
\end{table}
We conduct a detailed ablation study on the RAW-to-RAW mapping dataset~\cite{afifi2021semi} to assess the individual and cumulative impact of key components in \Mname. 
As shown in~\cref{tab:ablation}, our baseline model achieves reasonable performance, with an average MAE of $0.0221$, PSNR of $32.37$ dB, and SSIM of $0.8599$ across both translation directions. 
Notably, performance is asymmetric: Samsung-S9$\rightarrow$iPhone-X translation performs better across all metrics, indicating domain-specific variation in complexity.
Adding SANM leads to a substantial performance boost, particularly in the Samsung-S9$\rightarrow$iPhone-X direction. 
MAE drops to $0.0210$, while PSNR improves to $32.71$ dB, and SSIM rises to $0.8682$. 
This suggests that explicit modeling of sensor-specific noise profiles contributes significantly to image fidelity and structural consistency.
Incorporating MS-LKA alongside SANM further improves the overall translation quality. 
The average PSNR reaches $33.20$ dB, and the MAE is reduced to $0.0195$. 
These gains reflect MS-LKA’s ability to capture long-range dependencies and contextual structures, which are particularly beneficial in sensor-specific color or tone transitions that occur in challenging RAW domains.
Finally, adding an explicit consistency SSIM loss term yields the best overall performance. 
The average MAE drops to $0.0185$, PSNR improves to $33.60$ dB, and SSIM reaches $0.8806$. 
This confirms that directly optimizing for perceptual similarity complements the pixel-level losses.

\cref{tab:multi_domain_scaling} presents a comparative study of \mname against recent SOTA models, as the number of camera domains increases. 
Across all domain configurations, \mname achieves superior accuracy with significantly fewer parameters and lower training cost than UVCGAN, and matches or exceeds the performance of Xie~\etal despite requiring $2-5\times$ fewer training iterations. 
In the 3-domain case, \mname achieves the best performance with only 58.7M parameters, 1/3 the size of UVCGAN. 
Notably, while Xie~\etal has the smallest parameter count in this setting, it requires longer training ($266$K iterations), and still underperforms \mname in all metrics.
As the number of domains increases, \mname exhibits remarkable scalability. 
When moving from 3 to 5 domains, UVCGAN’s parameter count scales linearly (186M$\rightarrow$620M) and training time grows proportionally. 
In contrast, \mname maintains a nearly constant model size ($\sim$58.7M) and training iteration budget (180K), demonstrating strong efficiency and parameter-sharing capability. 
At 5 domains, \mname still outperforms the baselines in PSNR, SSIM, and MAE, while achieving the second-best KL divergence.
These results validate \Mname’s effectiveness in maintaining high quality, while being significantly more scalable and training-efficient than prior models. 

\noindent More results are discussed in the supplementary material.

\section{Related Work}
\label{eq:related_work}
\subsection{RAW-to-RAW Translation}
The objective of the RAW2RAW translation is to establish a mapping function $f$ capable of accurately translating RAW images from Camera A’s RAW space to Camera B’s RAW  space, accommodating diverse scenes and lighting conditions. 
In brief, the mapping function can be expressed as $I^B=f(I^A)$, where $I^A$ and $I^B$ denote the packed RAW images $[r, g_r, g_b, g]$ captured by camera A and camera B, respectively.
Work \cite{nguyen2014raw, kim2012new} proposed that by employing a quadratic transformation, the mapping function can be approximated by $I^B\approx I^A_{qt}T_{qt}$, where $I^A_{qt}=[r^2, g_r^2,\cdots,r\times g_r,g_r\times b,\cdots,r,g_r,g_b,g]$, and $T_qt\in\mathbb{R}^{14\times4}$ is a transformation matrix. 
Work \cite{afifi2021semi} was the first to
introduce neural networks into the RAW task; they placed a standard color chart in the scene and captured RAW images using two cameras. 
Subsequently, they estimated $T_{qt}$ by minimizing the color differences between the two color charts. 
They applied this transformation to the RAW images and obtained pixel-level paired data. 
They then utilized a neural network for semi-supervised training.
Although the trained model can accomplish RAW2RAW translation without having to capture paired images again, creating a paired dataset for training is still inevitable.
The following work \cite{perevozchikov2024rawformer, xie2024generalizing} released the requirement of using paired images for training. 
Rawformer \cite{perevozchikov2024rawformer} proposed a transformer-based encoder-decoder model to implement fully unsupervised learning.
They introduced contextual-scale aware downsampler and upsampler blocks that efficiently summarize the local-global contextual details in mixed scale
representations.
Concurrent work \cite{xie2024generalizing} proposed a color space predictor to predict the space transformation parameters in a patch-wise manner, which accurately performs transformation and flexibly manages complex lighting conditions.
Work \cite{nikonorov2025color} leveraged the spline capabilities of Kolmogorov-Arnold Networks \cite{liu2024kan} to model the color matching between source and target distributions. 
They developed a hypernetwork that generates spatially varying weights to control the nonlinear splines of a KAN, enabling accurate matching.
\subsection{Unpaired Image Translation}
Unpaired image translation aims to map images from the domain $I^A$ to $I^B$ without ground truth.
CycleGAN \cite{zhu2017unpaired} utilized a cycle-consistency loss and identity loss to bridge domain gaps.
UVCGAN \cite{torbunov2023uvcgan} introduced a pixel-wise transformer into the CycleGAN framework, while Swin-UNIT \cite{li2023swin} addressed performance issues caused by high-resolution images by introducing swin-transformer block.
StarGAN series \cite{choi2018stargan, choi2020stargan} extended the task from two-domain transformation to multi-domain.
StarGAN \cite{choi2018stargan} proposed a novel GAN that learns the mappings among multiple domains using only a single generator and a discriminator by adding a mask vector to the domain label.
StarGANv2 \cite{choi2020stargan} introduced a style encoder. 
The style encoder learns to extract the style code from a reference image from the target domain.
Given the style codes, the generator learns to synthesize images over multiple domains.
\vspace{-1em}
\section{Conclusion}
\label{sec:conclusion}
We presented \Mname, a novel and unified framework for multi-domain RAW image translation, capable of performing one-to-many and many-to-many mappings across diverse camera sensor domains using a single model. 
By explicitly modeling signal-dependent noise and enhancing global context perception, \mname significantly improves the fidelity and generalizability of RAW2RAW translations.
To support standardized benchmarking, we also introduced \Dname, the first dataset specifically curated for multi-domain RAW translation, including both paired and unpaired RAW captures from five diverse camera sensors under various real-world conditions.
Extensive experiments on existing datasets and \dname validate the advantages of \Mname. 
\mname achieves superior performance in terms of quality and scalability while maintaining a compact parameter footprint.
\section*{Acknowledgement}
This work was supported in part by the DARPA Young Faculty Award, the National Science Foundation (NSF) under Grants \#2431561,  \#2127780, \#2319198, \#2321840, \#2312517, and \#2235472, the Semiconductor Research Corporation (SRC), the Office of Naval Research through the Young Investigator Program Award and Grants \#N00014-21-1-2225 and \#N00014-22-1-2067, Army Research Office Grant \#W911NF2410360, and DARPA under Support Agreement No. USMA 23004. Additionally, support was provided by the Air Force Office of Scientific Research under Award \#FA9550-22-1-0253, along with generous gifts from Xilinx and Cisco.

{
    \small
    \bibliographystyle{ieeenat_fullname}
    \bibliography{main}
}


\clearpage

\maketitlesupplementary

\section{Dataset Details}
\label{sec:dataset_details}

This section provides detailed descriptions of the datasets used in our study.
We first introduce the \textbf{RAW-to-RAW Mapping Dataset}~\cite{afifi2021semi}, which serves as a benchmark for cross-sensor RAW color mapping.
Then, we present our newly collected \textbf{Multi-Domain RAW (MDRAW)} Dataset, designed for large-scale evaluation of multi-sensor RAW domain adaptation and alignment.

\subsection{RAW-to-RAW Mapping Dataset}
The RAW-to-RAW Mapping Dataset introduced by Afifi and Abuolaim~\cite{afifi2021semi} provides a benchmark for studying cross-sensor color-space mappings between different camera devices.
It contains RAW images captured by two smartphone cameras---\textit{Samsung Galaxy S9} and \textit{Apple iPhone X}---and includes both unpaired and paired subsets.

The unpaired set consists of \textbf{392 RAW images} (196 per camera) captured under various scenes and illumination conditions. 
The dataset also provides \textbf{115 paired testing images} and a small \textbf{anchor set} of 22 paired images, which were used for semi-supervised training and evaluation.

Paired samples were generated using a \textbf{color-chart-based polynomial calibration} between the two camera RAW spaces. 
For each scene, corresponding patches were manually selected from chart regions and homogeneous areas to compute the mapping matrix, ensuring more robust color alignment.

All images are provided in \textbf{DNG format} with black-level subtraction and normalization applied. 
Metadata such as black/white levels and color matrices are included. 
The dataset supports supervised, unsupervised, and semi-supervised RAW-domain learning.

\subsection{Multi-Domain RAW (MDRAW) Dataset}
As one of our main contributions, we introduce \textbf{Multi-Domain RAW (MDRAW)}, a new dataset of RAW images captured by multiple cameras with distinct sensor characteristics. 
It is designed to facilitate research on cross-sensor RAW image processing, domain adaptation, and universal RAW representation learning.

MDRAW contains RAW images captured by \textbf{five cameras} with different sensor designs and optics: 
\textit{Samsung Galaxy S23 Ultra}, \textit{Huawei P30}, \textit{iPhone 13 Pro}, \textit{Nikon Z5}, and \textit{Canon EOS Rebel T6}. 
The dataset includes both \textbf{unpaired} and \textbf{paired} RAW sets under diverse scenes and illumination conditions, along with their corresponding \textbf{sRGB references}.

For each camera, unpaired images were collected independently across indoor and outdoor environments, while paired data were captured under controlled conditions for cross-domain alignment. 
To construct a \textbf{pixel-level evaluation benchmark}, we further extracted spatially aligned patch pairs between images of the same scene taken by different devices using an extended \textbf{LoFTR-based correspondence pipeline} that combines dense matching, geometric verification, and patch-level synchronization.

All RAW files are stored in \textbf{DNG format} with corresponding metadata. 
\cref{tab:camera_sensors} summarizes the camera sensors and data statistics, and \cref{fig:dataset_visualization} visualizes representative samples across devices.

\section{MDRAW Construction via Cross-Domain LoFTR Matching}
\label{sec:mdraw_matching}

To quantitatively evaluate cross-domain RAW2RAW translation at the pixel level, we construct a high-quality benchmark dataset of paired RAW patches using real images captured by five distinct cameras: Huawei, Nikon, iPhone, Samsung, and Canon. Our goal is to extract spatially aligned patch pairs from images of the same scene taken by different devices, even in the presence of significant domain gaps in resolution, noise, exposure, and white balance.

\setcounter{subsection}{0}
\subsection{Cross-Domain Patch Pairing Pipeline}
\label{subsec:construction_pipeline}
Since no ground truth pixel-level alignment exists between images from different sensors, we adopt a feature-based correspondence approach. Specifically, we extend the LoFTR framework~\cite{sun2021loftr} to operate across domains, using a multi-stage pipeline that combines dense matching, geometric verification, spatial filtering, and patch-level synchronization.
We briefly outline the core stages below:

\noindent \textbf{Preprocessing and Normalization.}
Each RAW image is linearized by subtracting the camera-specific black level and normalizing by its dynamic range (white level $-$ black level). The Bayer data is then packed into a 4$-$channel RGGB format and spatially normalized via center-crop and downsampling to a fixed resolution. All images are rotated or flipped to ensure consistent spatial orientation across cameras.

\noindent \textbf{LoFTR-based Feature Matching.}
We convert normalized RGGB images to pseudo-RGB representations and extract grayscale versions for matching. We use LoFTR~\cite{sun2021loftr}, a detector-free dense matching method, to establish correspondences between grayscale image pairs. LoFTR excels in handling texture-poor regions and illumination inconsistencies, which are common issues in RAW domains.

\noindent \textbf{Geometric Verification.}
Candidate matches are filtered via RANSAC~\cite{fischler1981random} using homography estimation. Matches that are geometrically consistent form a reliable set of inliers for cropping.

\noindent \textbf{Spatial Non-Maximum Suppression (NMS).}
To ensure spatial diversity and avoid redundancy, we apply NMS on the match locations using a minimum center-to-center distance threshold. This encourages the patch pairs to cover various regions and scene content.

\noindent \textbf{Synchronized Cropping and Patch Extraction.}
For each retained match, we extract a 256×256 RGGB patch. Cropping is performed such that the patches are geometrically aligned at the matched keypoints, with optional correction for small shifts due to boundary constraints.


\section{Evaluation Metrics}
\label{sec:evaluation_metrics}
In this section, we provide detailed formulations of the four quantitative metrics used in our evaluation: 
Mean Absolute Error (MAE), 
Peak Signal-to-Noise Ratio (PSNR), 
Structural Similarity Index Measure (SSIM), 
and symmetric histogram-based KL divergence (KL Divergence). 
All metrics are computed on the denormalized RAW images in the linear intensity domain.

\subsection{Mean Absolute Error (MAE)}
MAE quantifies the average absolute difference between the predicted RAW image $I_p$ and the reference RAW image $I_r$:
\begin{equation}
\mathrm{MAE} = \frac{1}{N} \sum_{i=1}^{N} | I_p(i) - I_r(i) | ,
\end{equation}
where $N$ is the total number of pixels. 
As RAW data reside in the linear intensity domain, MAE directly reflects radiometric consistency and low-level reconstruction accuracy. 
A lower MAE indicates better preservation of RAW signal characteristics.

\subsection{Peak Signal-to-Noise Ratio (PSNR)}
PSNR measures the ratio between the maximum possible signal and the distortion noise, derived from the mean squared error (MSE):
\begin{align}
\mathrm{MSE} &= \frac{1}{N} \sum_{i=1}^{N} ( I_p(i) - I_r(i) )^2 ,\\
\mathrm{PSNR} &= 10 \log_{10} \left( \frac{L^2}{\mathrm{MSE}} \right) ,
\end{align}
where $L$ denotes the maximum possible pixel value (set to 1.0 after normalization). 
PSNR values range from 0 to $\infty$, with $\infty$ indicating perfect similarity, implying no discernible difference between the compared images.

\subsection{Structural Similarity Index Measure (SSIM)}
SSIM assesses perceptual image quality by comparing local luminance, contrast, and structural components between $I_p$ and $I_r$:
\begin{equation}
\mathrm{SSIM}(I_p,I_r) =
\frac{(2\mu_p \mu_r + C_1)(2\sigma_{pr} + C_2)}
     {(\mu_p^2 + \mu_r^2 + C_1)(\sigma_p^2 + \sigma_r^2 + C_2)} ,
\end{equation}
where $\mu_p, \mu_r$ are local means, $\sigma_p^2, \sigma_r^2$ are variances, and $\sigma_{pr}$ is the covariance. 
Constants $C_1$ and $C_2$ stabilize the division. 
SSIM varies from 0 to 1, where a value of 0 indicates no structural similarity between the images, and 1 denotes identical local structures.

\subsection{Symmetric KL Divergence (KL Divergence)}
We compute a symmetric, histogram-based Kullback--Leibler divergence to evaluate distributional similarity between $I_p$ and $I_r$:
\begin{align}
D_{\mathrm{KL}}(p\|q) &= \sum_{i} p_i \log \frac{p_i}{q_i} ,\\
D_{\mathrm{sym}}(p,q) &= \tfrac{1}{2} \big( D_{\mathrm{KL}}(p\|q) + D_{\mathrm{KL}}(q\|p) \big) .
\end{align}
The probability distributions $p$ and $q$ are estimated from per-channel intensity histograms (256 bins, range [0,1]), with small-value regularization $\varepsilon$ added to ensure numerical stability. 
This metric captures global statistical differences between RAW image distributions across sensors.

Together, these four metrics provide complementary perspectives on both pixel-level reconstruction accuracy and overall distributional alignment in the RAW domain.

\section{More Results}
\label{sec:more_results}

\begin{figure}[ht]
  \centering
  \includegraphics[width=\linewidth]{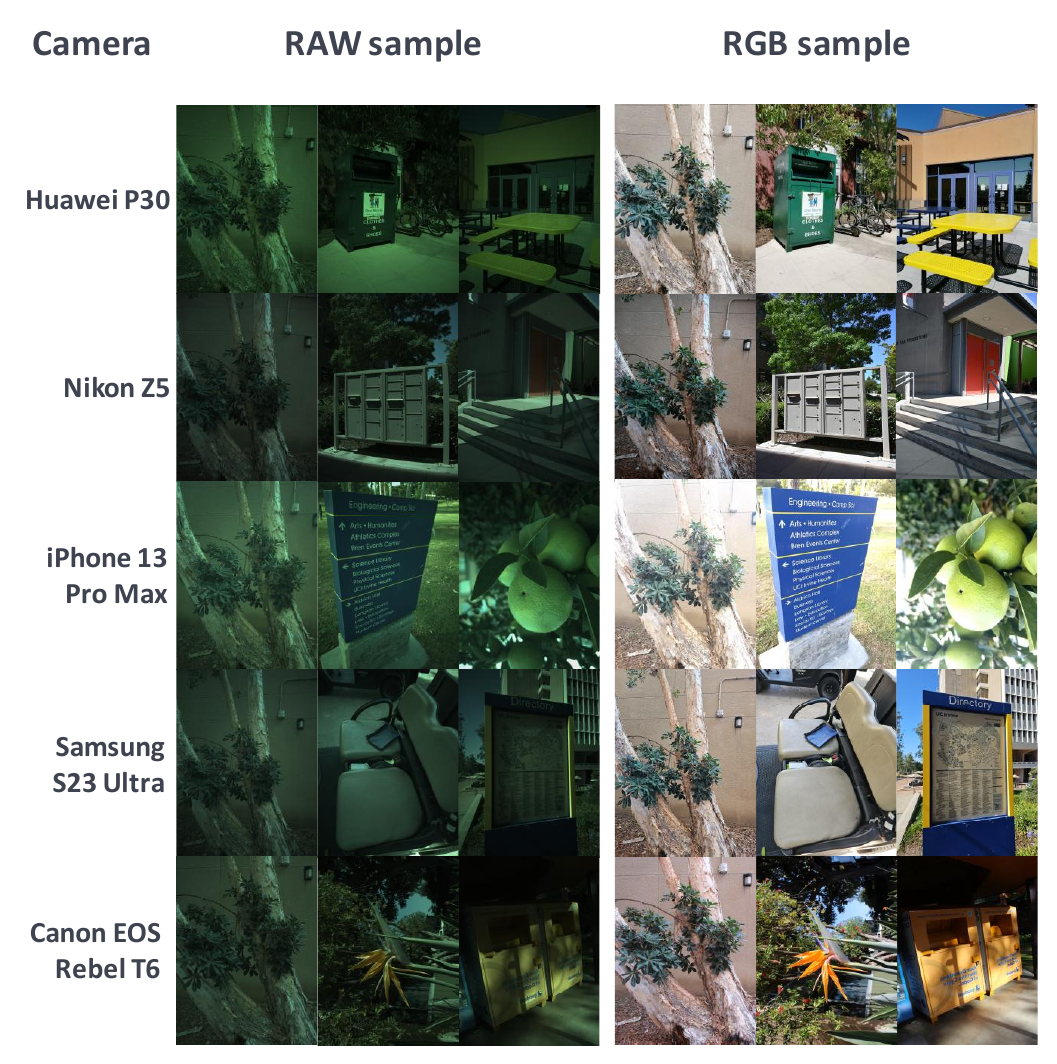}
  \vspace{-2em}
   \caption{\textbf{Visualization of more captured images in \Dname.}}
    \label{fig:sup_vis}
    \vspace{-1em}
\end{figure}
\cref{fig:sup_vis} illustrates the diversity and cross-domain characteristics of the proposed \Dname dataset. 
Each row corresponds to a specific camera sensor, including Huawei P30, Nikon Z5, iPhone 13 Pro Max, Samsung S23 Ultra, and Canon EOS Rebel T6, showcasing both RAW and their corresponding RGB images. 
The RAW samples reveal substantial variations in tone, exposure, and noise characteristics across devices, even under similar scene conditions. 
These differences stem from the distinct spectral response functions, sensor noise patterns, and in-camera processing pipelines of each sensor. 
In contrast, the RGB samples appear more visually consistent due to the influence of proprietary ISP modules. 
This highlights the challenge and necessity of direct RAW2RAW translation to enable fair cross-domain vision tasks without relying on device-specific ISP outputs. 
\cref{fig:sup_vis} also underscores the importance of building datasets like \Dname to support learning models capable of generalizing across multiple RAW domains.

\begin{figure*}[ht]
    \centering
    \includegraphics[width=.9\linewidth]{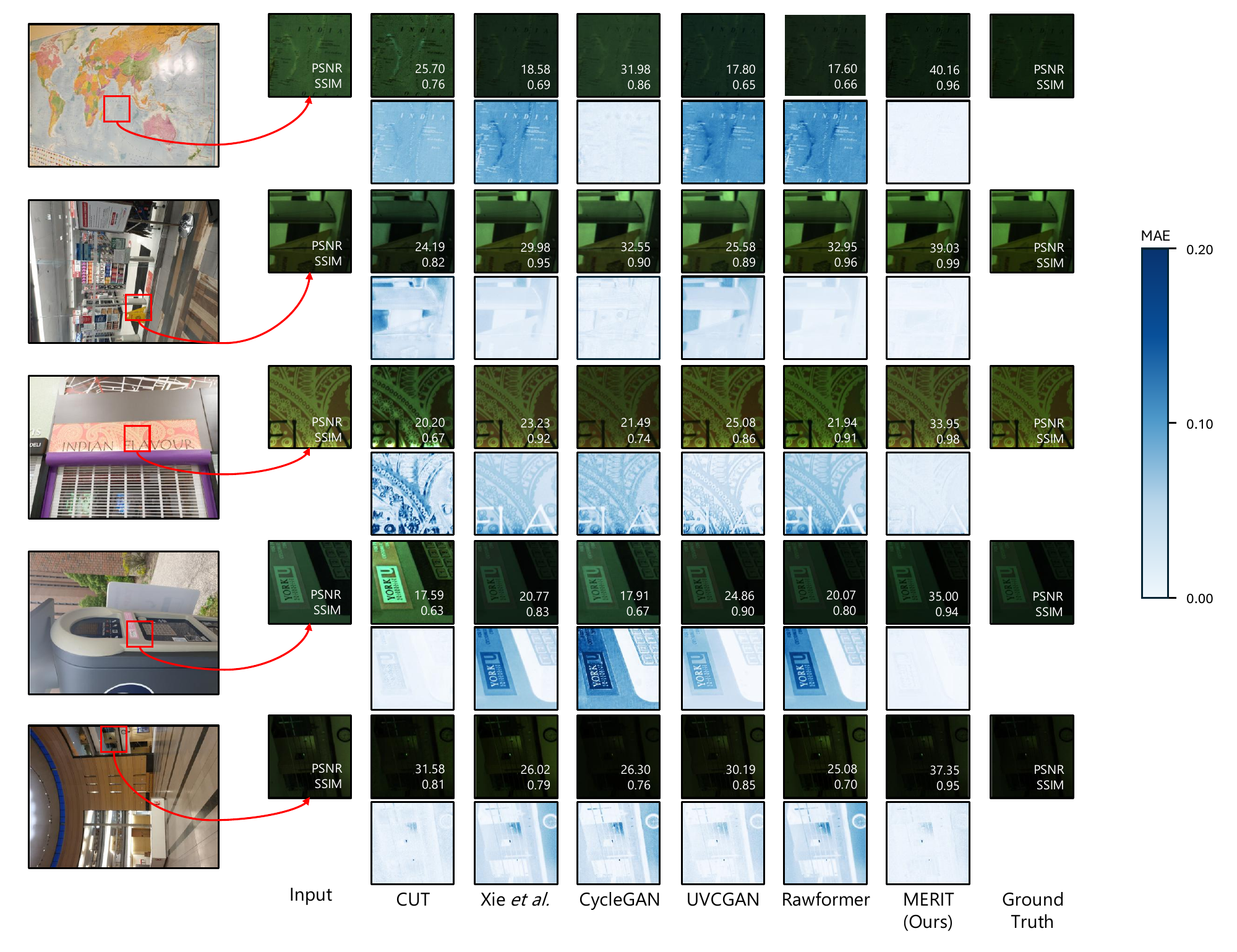}
    \caption{More qualitative results on the RAW-to-RAW mapping dataset.}
    \label{fig:sup_quali}
\end{figure*}

\cref{fig:sup_vis} provides a visual comparison of RAW-to-RAW translation results across five challenging real-world scenes, highlighting the performance of different methods. 
Each row corresponds to a different scene, with the source image on the left, followed by translated outputs, and the ground truth on the right. 
For each method, the translated patch and its corresponding absolute error map (visualized via MAE) are shown, along with PSNR and SSIM metrics.

Across all examples, \mname consistently produces translated outputs that are both visually faithful and quantitatively superior. 
In the first scene (world map), \mname accurately reconstructs the fine textural patterns and subtle gradients with the highest PSNR (40.16) and SSIM (0.96), significantly outperforming other methods that either blur the details or introduce artifacts. 
In the second scene (indoor poster), \mname achieves a near-perfect reconstruction (PSNR: 39.03, SSIM: 0.99), preserving text clarity and illumination distribution that others fail to retain.
In the third and fourth scenes, characterized by repetitive textures and sharp edges, \mname demonstrates superior structure preservation. 
Competing methods such as CycleGAN and Rawformer often produce distorted patterns or overly smoothed outputs, as evident in the high error maps. 
Conversely, \mname maintains edge consistency and semantic accuracy, with notably lower MAE.
Finally, in the fifth example (indoor hallway), \mname again achieves the best perceptual quality and the most faithful reconstruction of lighting and geometry. 
The overall results demonstrate \Mname’s strong generalizability across diverse domains and scenes, as well as its ability to balance perceptual quality and pixel-level fidelity more effectively than prior methods.

\begin{table}[t]
\centering
\resizebox{\linewidth}{!}{%
\begin{tabular}{c|c|c|c|c|c|c|c}
\toprule\toprule
\textbf{No. of Domains} & \textbf{Method} & \textbf{Parameters (M)} & \makecell{\textbf{Training Iteration (K)} / \\ \textbf{Training Time (Minutes)}} & \textbf{Avg MAE}$\downarrow$ & \textbf{Avg PSNR}$\uparrow$ & \textbf{Avg SSIM}$\uparrow$ & \textbf{Avg KL Div}$\downarrow$ \\
\midrule\midrule
\multirow{4}{*}{3}
  & UVCGAN~\cite{torbunov2023uvcgan}         & 186 &228 /  1955  &0.039  &\underline{28.03}  &0.72  &2.72  \\
  & Xie \etal~\cite{xie2024generalizing}      &\cellcolor{gray!20}\textbf{33.6} &\cellcolor{gray!20}266 / 1338  &\cellcolor{gray!20}\underline{0.038}  &\cellcolor{gray!20}27.82  &\cellcolor{gray!20}\underline{0.73}  &\cellcolor{gray!20}\underline{2.33}  \\
  & \textbf{MERIT (Ours)} &\cellcolor{gray!50}\underline{58.7}  &\cellcolor{gray!50}120 / 1400&\cellcolor{gray!50}\textbf{0.031} &\cellcolor{gray!50}\textbf{29.60}  &\cellcolor{gray!50}\textbf{0.76}  &\cellcolor{gray!50}\textbf{1.76}  \\
\midrule
\multirow{4}{*}{4}
  & UVCGAN~\cite{torbunov2023uvcgan}         & 372 &476 / 3956  &0.038  &27.85  &\underline{0.72}  &2.66  \\
  & Xie \etal~\cite{xie2024generalizing}      &\cellcolor{gray!20}\underline{67.2}  &\cellcolor{gray!20}507 / 2055  &\cellcolor{gray!20}\underline{0.037 } &\cellcolor{gray!20}\underline{28.31}  &\cellcolor{gray!20}\underline{0.72}  &\cellcolor{gray!20}\underline{2.35}  \\
  & \textbf{MERIT (Ours)} &\cellcolor{gray!50}\textbf{58.7}  &\cellcolor{gray!50}165 / 1925 &\cellcolor{gray!50}\textbf{0.033}  &\cellcolor{gray!50}\textbf{29.08}  &\cellcolor{gray!50}\textbf{0.74}  &\cellcolor{gray!50}\textbf{2.07}  \\
\midrule
\multirow{4}{*}{5}
  & UVCGAN~\cite{torbunov2023uvcgan}         & 620 &748 / 6581  &0.038  &28.10  &0.72  &2.48  \\
  & Xie \etal~\cite{xie2024generalizing}      &\cellcolor{gray!20}\underline{112}  &\cellcolor{gray!20}901 / 3428 &\cellcolor{gray!20}\underline{0.037}  &\cellcolor{gray!20}\underline{28.23}  &\cellcolor{gray!20}\underline{0.73}  &\cellcolor{gray!20}\textbf{2.28}  \\
  & \textbf{MERIT (Ours)} &\cellcolor{gray!50}\textbf{58.9}  &\cellcolor{gray!50}180 / 2100
  &\cellcolor{gray!50}\textbf{0.036}  &\cellcolor{gray!50}\textbf{28.42}  &\cellcolor{gray!50}\textbf{0.74}  &\cellcolor{gray!50}\underline{2.30}  \\
\bottomrule\bottomrule
\end{tabular}
}
\vspace{-1em}
\caption{\textbf{Comparison of different models under increasing numbers of domains.} \mname demonstrates both competitive performance and efficiency.}
\label{tab:multi_domain_scaling}
\end{table}

\begin{table}[h!]
\centering
\resizebox{\linewidth}{!}{
\begin{tabular}{c|c|c|c|c}
\toprule\toprule
\textbf{\diagbox{Source}{Target}} & \textbf{Samsung} & \textbf{Huawei} & \textbf{iPhone} & \textbf{Nikon} \\
\midrule
\textbf{Samsung} & -- & 
\makecell{ 
0.025 / 31.23 / 0.77 / 1.33
} 
& \makecell{ 
0.036 / 29.02 / 0.76 / 1.78
}
& \makecell{ 
0.035 / 28.45 / 0.72 / 2.11
} 

\\
\midrule
\textbf{Huawei}  
& \makecell{ 
0.029 / 30.11 / 0.74 / 1.70
}
& -- 
&   \makecell{ 
0.033 / 29.12 / 0.77 / 2.34
}
&   \makecell{ 
0.032 / 29.42 / 0.74 / 1.95
}

\\
\midrule
\textbf{iPhone}  
& \makecell{ 
0.035 / 28.88 / 0.73 / 1.56
} 
&  \makecell{ 
 0.030 / 29.69 / 0.76 / 1.95
} 
& -- 
&  \makecell{ 
0.039 / 27.52 / 0.71 / 2.38
}

\\
\midrule
\textbf{Nikon}   
& \makecell{ 
0.040 / 27.22 / 0.68 / 3.00
}
& \makecell{ 
0.034 / 28.71 / 0.73 / 2.69
}
&  \makecell{ 
0.043 / 26.51 / 0.71 / 3.34
}
& -- 

\\
\bottomrule\bottomrule
\end{tabular}
}
\caption{\textbf{4$\times$4 Cross-domain RAW2RAW translation results on \Dname.}
Each cell reports results for translation from a source domain (column) to a target domain (row).
Each line contains four metrics in the order of \textbf{MAE}($\downarrow$)/ \textbf{PSNR}($\uparrow$)/ \textbf{SSIM} ($\uparrow$)/ \textbf{KL Divergence}($\downarrow$). }
\label{tab:cross_domain_4}
\end{table}

\begin{table}[h!]
\centering
\resizebox{\linewidth}{!}{
\begin{tabular}{c|c|c|c}
\toprule\toprule
\textbf{\diagbox{Source}{Target}} & \textbf{Samsung} & \textbf{Huawei} & \textbf{iPhone} \\
\midrule
\textbf{Samsung} & -- & 
\makecell{ 
0.025 / 31.24 / 0.77 / 1.36
} 
& \makecell{ 
0.036 / 29.02 / 0.76 / 1.63
}
 
\\
\midrule
\textbf{Huawei}  
& \makecell{ 
0.029 / 30.04 / 0.75 / 1.48
}
& -- 
&   \makecell{ 
0.034 / 28.96 / 0.77 / 2.27
}

\\
\midrule
\textbf{iPhone}  
& \makecell{ 
0.035 / 28.53 / 0.72 / 1.65
} 
&  \makecell{ 
 0.036 / 29.02 / 0.76 / 1.63
} 
& -- 

\\
\bottomrule\bottomrule
\end{tabular}
}
\caption{\textbf{3$\times$3 Cross-domain RAW2RAW translation results on \Dname.}
Each cell reports results for translation from a source domain (column) to a target domain (row).
Each line contains four metrics in the order of \textbf{MAE}($\downarrow$)/ \textbf{PSNR}($\uparrow$)/ \textbf{SSIM} ($\uparrow$)/ \textbf{KL Divergence}($\downarrow$). }
\label{tab:cross_domain_3}
\end{table}

To evaluate the generalization and scalability of our proposed \mname framework, we compare its cross-domain translation performance when trained on 3 domains (Samsung, Huawei, iPhone) versus 4 domains (adding Nikon). 
As shown in \cref{tab:cross_domain_3} and \cref{tab:cross_domain_4}, \mname consistently maintains strong translation performance across both settings. 
Notably, the average PSNR across all 3$\times$3 transfers remains high (e.g., 30.04$\rightarrow$31.24 for Samsung$\rightarrow$Huawei), with marginal variation from the corresponding 4$\times$4 results. 
For instance, the translation from Samsung to Huawei achieves 31.24 dB PSNR in both settings, while Huawei to iPhone reaches 29.12 dB in the 4-domain setup and 28.96 dB in the 3-domain setup. 
Similarly, SSIM values remain stable across settings. 
These results indicate that \mname generalizes well to varying numbers of domains without significant degradation. 
Furthermore, the stable performance with the added domain (Nikon) demonstrates \mname’s scalability: the unified model scales gracefully to more diverse camera domains while maintaining high-quality translation, supporting its deployment in real-world multi-sensor environments.

\begin{table}[h!]
\centering
\resizebox{.9\linewidth}{!}{
\begin{tabular}{c|c|c|c|c}
\toprule\toprule
\textbf{Loss Term} & \textbf{Setting} &
\textbf{MAE}~\(\downarrow\) &
\textbf{PSNR}~\(\uparrow\) &
\textbf{SSIM}~\(\uparrow\) \\
\midrule

\multirow{5}{*}{$\mathcal{L}_{noise}$}
& $\lambda = 0.25$ & 0.0203 & 32.81 & 0.08668 \\
& $\lambda = 0.75$ & 0.0199 & 33.27 & 0.8707 \\
& $\lambda = 1$     & \textbf{0.0185} & \textbf{33.60} & \textbf{0.8806} \\
& $\lambda = 1.5$   & 0.0195 & 33.24 & 0.8769 \\
& $\lambda = 5$     & 0.0212 & 32.36 & 0.8576 \\
\midrule

\multirow{5}{*}{$\mathcal{L}_{cycle-SSIM}$}
& $\lambda = 0.02$ & 0.0222 & 32.21 & 0.8560 \\
& $\lambda = 0.05$ & 0.0200 & 33.18 & 0.8735 \\
& $\lambda = 0.08$ & 0.0211 & 32.54 & 0.8683 \\
& $\lambda = 0.1$  & \textbf{0.0185} & \textbf{33.60} & \textbf{0.8806} \\
& $\lambda = 0.2$  & 0.0192 & 33.18 & 0.8695 \\
\bottomrule\bottomrule
\end{tabular}
}
\caption{\textbf{Ablation study on loss weight sensitivity.}
We evaluate the impact of loss weighting coefficients $\lambda$ for the proposed noise modeling loss $\mathcal{L}_{\text{noise}}$ and the cycle consistency SSIM loss $\mathcal{L}_{\text{cycle-SSIM}}$.
Each cell reports average performance on the RAW-to-RAW mapping dataset.
Best results for each loss term are highlighted in bold.}
\label{tab:loss_hyperparam}
\end{table}

We conduct a thorough ablation study to examine the sensitivity of \mname to the weighting of its two proposed loss terms, $\mathcal{L}_{\text{noise}}$ and $\mathcal{L}_{\text{cycle-SSIM}}$, as shown in \cref{tab:loss_hyperparam}.
For $\mathcal{L}_{\text{noise}}$, performance improves steadily as the weight $\lambda$ increases from 0.25 to 1.0, reaching optimal results at $\lambda = 1$ with a PSNR of 33.60, SSIM of 0.8806, and MAE of 0.0185. 
Beyond this point, further increasing the weight degrades performance, suggesting that overly emphasizing noise alignment may compromise semantic fidelity or overall reconstruction quality.
A similar trend is observed for $\mathcal{L}_{\text{cycle-SSIM}}$. 
The optimal setting is again $\lambda = 0.1$, producing identical peak performance across all three metrics. 
Lower weights underutilize the semantic supervision provided by SSIM, while larger values likely over-constrain the reconstruction, leading to degraded perceptual quality.
Overall, both proposed loss terms contribute positively to performance when appropriately weighted, and the consistent peak at $\lambda = 1$ and $\lambda = 0.1$, respectively, highlights the effectiveness and stability of these components in the \Mname.


\end{document}